\documentclass[letterpaper]{article} 
\usepackage{aaai25}  
\usepackage{times}  
\usepackage{helvet}  
\usepackage{courier}  
\usepackage[hyphens]{url}  
\usepackage{graphicx} 
\usepackage{natbib}  
\usepackage{caption} 
\usepackage{graphicx}
\usepackage{amsmath}  
\usepackage{amssymb}  
\usepackage{amsfonts} 
\usepackage{xcolor}

\usepackage{amsmath}
\usepackage{array}
\usepackage{booktabs}
\usepackage{multirow}

\usepackage{algorithm}
\usepackage{algpseudocode}  
\usepackage{amsmath}

\usepackage{newfloat}
\usepackage{listings}
\DeclareCaptionStyle{ruled}{labelfont=normalfont,labelsep=colon,strut=off} 
\floatstyle{ruled}
\newfloat{listing}{tb}{lst}{}
\floatname{listing}{Listing}
\title{Activation Space Selectable Kolmogorov-Arnold Networks}
\author{
    Zhuoqin Yang\textsuperscript{\rm 1*}, Jiansong Zhang\textsuperscript{\rm 2*}, Xiaoling Luo\textsuperscript{\rm 2},
    Zheng Lu\textsuperscript{\rm 1\dag}, Linlin Shen\textsuperscript{\rm 2\dag}
}
\affiliations{
    \textsuperscript{\rm 1}School of Computer Science , University of Nottingham Ningbo China, Ningbo, China\\
    \textsuperscript{\rm 2} School of Computer Science \& Software Engineering, Shenzhen University, Shenzhen, China\\
    Emails: zhuoqin.yang@nottingham.edu.cn, jiansongzhang3111@gmail.com, xiaolingluoo@outlook.com, zheng.lu@nottingham.edu.cn, llshen@szu.edu.cn \\
     \textsuperscript{*}These authors contributed equally to this work. 
    \textsuperscript{\dag}Corresponding authors
}

\usepackage{bibentry}

\begin{document}

\maketitle

\begin{abstract}
The multilayer perceptron (MLP), a fundamental paradigm in current artificial intelligence, is widely applied in fields such as computer vision and natural language processing. However, the recently proposed Kolmogorov-Arnold Network (KAN), based on nonlinear additive connections, has been proven to achieve performance comparable to MLPs with significantly fewer parameters. Despite this potential, the use of a single activation function space results in reduced performance of KAN and related works across different tasks. To address this issue, we propose an activation space \textbf{S}electable \textbf{KAN} (\textbf{S-KAN}). S-KAN employs an adaptive strategy to choose the possible activation mode for data at each feedforward KAN node. Our approach outperforms baseline methods in seven representative function fitting tasks and significantly surpasses MLP methods with the same level of parameters. Furthermore, we extend the structure of S-KAN and propose an activation space selectable Convolutional KAN (S-ConvKAN), which achieves leading results on four general image classification datasets. Our method mitigates the performance variability of the original KAN across different tasks and demonstrates through extensive experiments that feedforward KANs with selectable activations can achieve or even exceed the performance of MLP-based methods. This work contributes to the understanding of the data-centric design of new AI paradigms and provides a foundational reference for innovations in KAN-based network architectures.

\end{abstract}

%

\section{Introduction}

Over an extended period, the consensus in designing machine learning paradigms based on neural networks has been to establish nonlinear mappings from basic linear spaces \cite{schetzen1985multilinear}. Whether in the field of computer vision or natural language processing, Multilayer Perceptrons (MLP) that combine linear and nonlinear parametrization, form the cornerstone of these domains. As a fundamental architecture in modern AI frameworks, MLPs have demonstrated exceptional benchmarks in traditional paradigm designs \cite{cheng2024multilinear}. Particularly in computer vision, paradigms represented by Convolutional Neural Networks (CNN) \cite{gu2018recent}, Spiking Neural Networks (SNN) \cite{ghosh2009spiking}, and Visual Attention \cite{guo2023visual} have stood up due to MLP's parametric and mnemonic capabilities. However, the introduction of Kolmogorov-Arnold Networks (KAN), based on the Kolmogorov-Arnold theorem, purposefully demonstrated that relying on nonlinear data summation can match or even surpass the solutions offered by traditional MLP-based approaches. KANs employ activation layers directly for data embedding, utilizing matching summation strategies for forward propagation. Mostly, operations involving linear parametric multiplications are discarded, with model learning achieved through the training of activation functions. The success of KANs has spurred exploration into the design of variable activation functions. 

The traditional paradigm of learning linear parameters from data is gradually shifting towards constructing patterns of nonlinear association. Since the proposal of KAN, a plethora of designs using well-known activation functions, such as ReLU \cite{qiu2024relu} and Chebyshev polynomials, for feature extraction, have emerged, specifically considering prior assumptions about domains like time series analysis and signal processing. However, the domain-specific applicability of these designs under specific paradigms casts a shadow on the broader adoption of the KAN paradigm. The issue of extending the use of KANs remains underexplored in these specific activation function designs and domain applications, as each KAN design is obscured by domain-specific challenges. This provokes a reflection: given that KAN is a structural foundation for learning from nonlinearity, how can we unify or simplify the selection of different KAN's paradigms?

\begin{figure*}[h]
\centering
\includegraphics[scale=0.9]{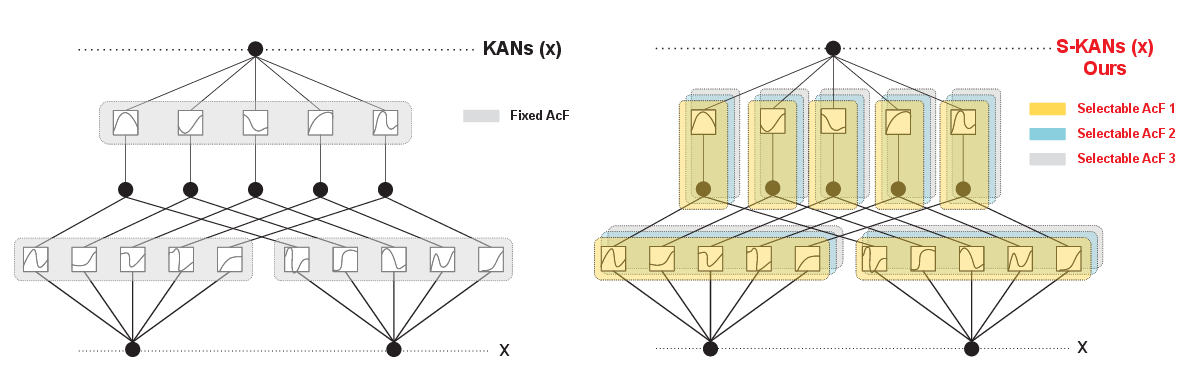}
\caption{KAN with fixed activation function layers (left) and the activation spaces Selectable KAN (S-KAN) proposed in this paper (right). Compared to the restricted activation function usage in the original KAN, S-KAN performs adaptive selection of activation functions for each data node, ensuring that the feedforward activation functions are the best choices under the selected strategy.}
\label{figg1}
\end{figure*}

In light of the aforementioned issues, this paper begins by examining the nonlinear characteristics and paradigm selection issues of KAN, proposing an adaptive activation space selection mechanism for nonlinear KAN data fitting, as shown in Figure \ref{figg1}. Specifically, the adaptive activation space encompasses the existing KAN designs, supplemented by linear adaptive parameters outside these nonlinear activation layers to compensate for the coarse designs derived from task understandings. It demonstrates good performance in multiple data fitting and computer vision tasks, without significantly increasing the number of parameters, surpassing those KAN paradigms specifically designed for particular domains. In summary, the contributions of this paper are as follows:
\begin{itemize}
\item This paper analyzes the deficiencies and shortcomings of existing KAN designs from the perspectives of linear data fitting and nonlinear activation, proposing S-KAN with an adaptive activation space selection that achieves nonlinear fitting purely from data, independent of any priors.
\item This paper conducts adaptive analysis and testing of the activation space for various data fitting tasks, demonstrating that the KAN paradigm with adaptive activation space selection, effectively compensates for the shortcomings of the original KAN designs and is suitable for classical machine learning tasks.
\item This paper improves existing CNN-KAN and proposes S-ConvKAN by utilizing nonlinear image feature extractor. It surpassed CNN with a similar number of parameters and other KAN-based networks in general computer vision tasks, achieving a leading level of performance.
\end{itemize}

\section{Related Work}
\subsection{MLP-based Paradigms}
Since the perceptron was proposed \cite{rosenblatt1958perceptron}, it has consistently played a significant role in the field of artificial intelligence \cite{krenn2023forecasting}. The perceptron aims to simulate the brain by using linear combinations and nonlinear activations to achieve parametric memory and learning \cite{carpenter1989neural,amit1989modeling}. In recent years, with the growing interest in artificial general intelligence (AGI) research \cite{kirillov2023segment}, MLP has shown great vitality in many fields \cite{pei2019towards,xu2021artificial,fei2022towards}. In the broad AI paradigm, almost all feature extractors have adopted the concept of the perceptron, such as CNN \cite{lecun2010convolutional,liu2022convnet,ding2022scaling}, SNN \cite{lindblad2005image,tavanaei2019deep,yao2023attention}, recurrent neural networks (RNN) \cite{graves2012long,zaremba2014recurrent,peng2023rwkv}, and Transformer networks \cite{vaswani2017attention,gu2023mamba}. Particularly noteworthy is that, under the premise of tokenized data understanding, it has become possible to use pure MLPs as feature extractors \cite{tolstikhin2021mlp, fusco2022pnlp, yu2022s2,ekambaram2023tsmixer}. From the perspective of tokenized data organization, fully connected MLPs can function in distinct regions, ultimately achieving their functionality through the fusion of these sparse representations \cite{lin2023ps}. MLPs form the foundation of current AI design, and understanding AI design through fully connected node operations has become a consensus \cite{jordan2015machine, chang2024survey}. However, the large number of parameters also limits the application of MLPs \cite{Lin_2024_CVPR}.
\subsection{Exploration on KANs}
The excessive number of parameters in MLPs has brought new feature computation methods \cite{li2020wavelet,li2021wavecnet,finder2024wavelet} into the research spotlight. KAN was proposed based on considerations of the nonlinear aspects of data. The learnability of nonlinear data fitting stems from the design of nonlinear activation functions with variable learnable parameters. In the original work by Liu \cite{liu2024kan}, B-spline functions were employed for data fitting, initially demonstrating extended KANs beyond the Kolmogorov-Arnold (K-A) theorem. These expanded KANs exhibit superior data fitting capabilities compared to MLPs, with minimal parameterization and enhanced interpretability. Subsequently, the development of KANs has progressed along two main paths. On one hand, the foundational designs of KAN paradigms have spurred extensive exploration around activation function design, inspired by the learnable activation space introduced by variable scale factors of B-spline functions. The BSRBF-KAN \cite{ta2024bsrbf}, integrates B-spline and radial basis functions to address stability and convergence issues in the original KAN framework. Additionally, Bozorgasl et al. proposed WAV-KAN \cite{bozorgasl2024wav}, a variable-scale KAN paradigm designed from wavelet transforms. They utilized discrete wavelet transforms to control scale factors for nonlinear data fitting, aiming to enhance the applicability of general KAN paradigms across tasks. Furthermore, efforts have been made to improve KAN computational efficiency, including the use of low-rank strategies \cite{li2024kolmogorovarnold} and GPUs \cite{torchkan} for accelerated computing. On the other hand, KAN has been practically applied as a foundational method replacing MLPs, in general tasks \cite{genet2024tkan,vaca2024kolmogorov,bresson2024kagnns}. Drokin initially combined CNNs with KANs \cite{drokin2024kolmogorovarnoldconvolutionsdesignprinciples}, employing reparameterization to enhance KAN's capabilities in image feature extraction. Dylan Bodner et al. further demonstrated the use of KAN strategies on each convolution operator, instead of traditional neural network methods, for feature computation \cite{bodner2024convolutional}. Their work effectively illustrates that KAN paradigms can serve as feature extractors in broad machine learning domains such as computer vision.
\section{Method}
\subsection{Primarily: Kolmogorov-Arnold Networks}
\subsubsection{Non-linear Learnable}

Different from the Multilayer Perceptron (MLP) structure based on fully connected neural networks, the Kolmogorov-Arnold Networks (KAN) are grounded in the Kolmogorov-Arnold theorem, utilizing nonlinear functions for data fitting. The nonlinear fitting design following the K-A theorem can be summarized as follows: for any continuous multivariate function \( f(x_1, x_2, \ldots, x_n) \), there exists a set of continuous univariate functions \( \phi_i \) and \( \phi_{ij} \) such that this multivariate function can be expressed as a linear combination of these univariate functions, as shown in Eq \ref{eq:ka}, where \(\phi (\cdot)\) represents the functional correspondence rule and \( \phi_{i} \) and \( \phi_{ij} \) are univariate functions.
\begin{equation}
f(x_1, x_2, \ldots, x_n) = \sum_{i=1}^{2n+1} \phi_i \left( \sum_{j=1}^{n} \phi_{ij} (x_j) \right)
\label{eq:ka}
\end{equation}

Inspired by the K-A theorem, \( \phi_i \) and \( \phi_{ij} \) are defined as learnable nonlinear data embedding functions, called \textit{Activation Functions (AcFs)}. These AcFs are applied with deep structures that extend beyond the number of connections specified by the K-A theorem. In this context, the learning ability of the system is maintained by adjusting the learnable parameters of the AcFs. Ultimately, the data undergoes relative positional summation and feedforward under the adaptive activation functions. 

\subsubsection{Activation Space in KANs}
\begin{figure}[h]
\centering
\includegraphics[scale=0.19]{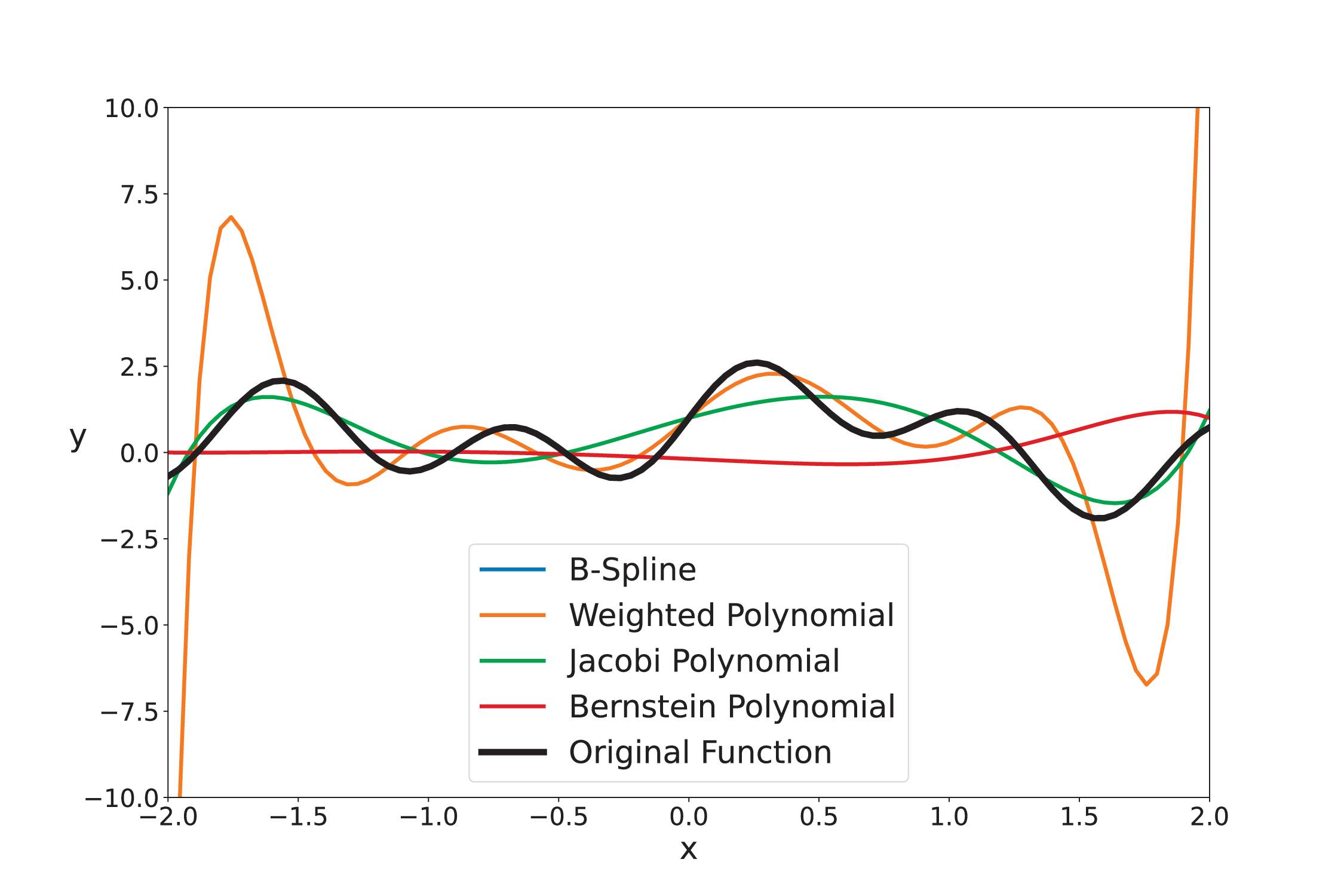}
\caption{In univariate function fitting (black), different fitting strategies (other colors) exhibit significant differences in performance at various values.}
\label{figg2}
\end{figure}

According to the Kolmogorov-Arnold theorem, theoretically, an infinite number of univariate functions can be used to decompose the fitting curve. However, due to resource constraints, most KANs utilize a finite number of data points or activation functions with adjustable ranges as basic computational units. This leads to differences when using various nonlinear fitting methods. As shown in Figure \ref{figg2}, we visualize the performance of different \( \phi_i \) in univariate function fitting. At different values, various known univariate nonlinear fitting functions exhibit discrepancies.

Ideally, we hope that KAN will demonstrate robust fitting characteristics at different fitting values. However, merely adjusting the learnable parameters of a single activation function is insufficient to achieve this. Therefore, we summarize the design of the known nonlinear function fitting factors used in KAN, into an activation space $\mathcal{F}$ with $m$ kinds of activation functions, as shown in Eqs \ref{eq:act_space1} and \ref{eq:act_space}. The designed KAN model should adaptively select those functional methods that beneficially reduce fitting errors during data learning. Rule-based activation path selection endows data fitting based on the KAN method with greater operability and can theoretically significantly improve the performance of existing KAN paradigms.

\begin{equation}
\label{eq:act_space1}
\mathcal{F} = \{ \textit{B-spline, RBF, Chebyshev \ldots} \}
\end{equation}

\begin{equation}
\Phi = \{ \phi^{(K_p)} : \mathbb{R} \to \mathbb{R} \mid K_p \in \mathcal{F}, p \in \{1, 2, \ldots, m\} \}
\label{eq:act_space}
\end{equation}
\begin{figure*}[h]
\centering
\includegraphics[scale=0.8]{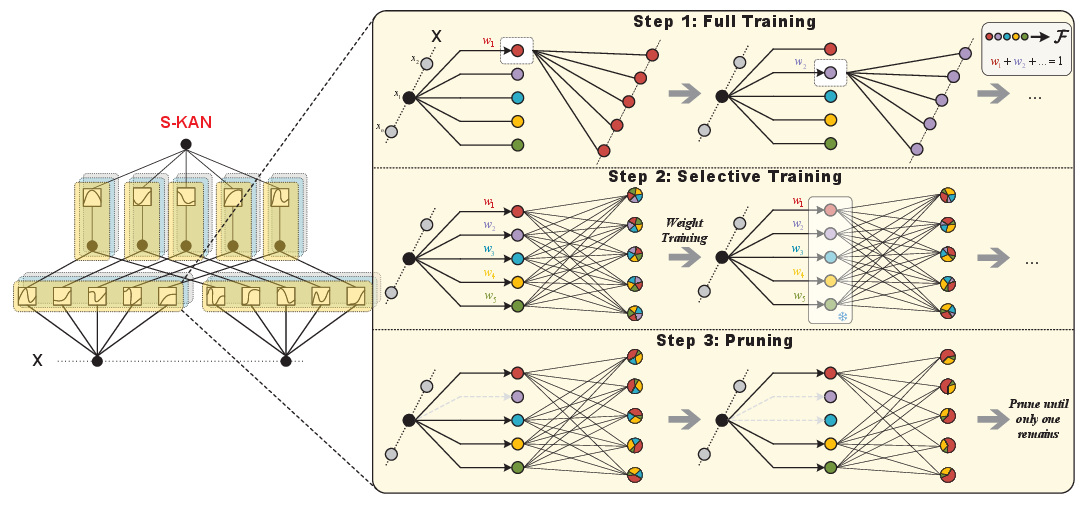}
\caption{The three-step training strategy in S-KAN. \textbf{Step 1: Full Training.} Each candidate nonlinear mapping method in the function pool initializes feedforward weights, which, together with the activation function scale factors, receive gradients from the cost function and are updated. \textbf{Step 2: Selective Training.} Freeze the activation function scale factors and train the selective weights to optimize the cost function. \textbf{Step 3: Pruning.} Prune the unimportant activation functions based on their weights. After pruning, the remaining weights are redistributed to ensure that they sum to 1.
}
\label{flow_fig}
\end{figure*}
\subsection{Selectable Kolmogorov-Arnold Networks(S-KAN)}
\subsubsection{Selectable Activation Space} Common function fitting methods, such as B-spline functions, Chebyshev polynomials, and wavelet transforms, are used to construct the activation function pool of the KAN. For the same type of polynomial or fitting function, we applied different orders or types of hyperparameters. Additionally, the activation functions used by existing methods designed for the KAN were also added to the activation function pool. S-KAN uses the activation functions in the pool to pre-train the samples and selects the most suitable activation function for each node in the KAN. In subsequent training, these selected activation functions are combined to achieve better results.

Specifically, in the initial stage, we will use a linear combination of all activation functions to complete the S-KAN prediction. Given the activation function pool $\mathcal{F}$, for each node in S-KAN, the initial prediction formula is:

\begin{equation}
\textit{Output$_{<node>}$}(x_i) = \sum_{p=1}^m w_p \cdot \phi_i^{(K_p)} (x_i)
\label{eq:combined}
\end{equation}
where \( w_{p} \), \( p \in \{1, 2, \ldots, m\} \), is the weight of the \( p \)-th activation function $AcF_{i}$. The initial weight of each activation function is uniformly set to \( w_{p} = \frac{1}{m} \), ensuring fair selection of each activation function.

\subsubsection{Efficient S-KAN Training}
The workflow of S-KAN is shown in Figure \ref{flow_fig}. The model undergoes a series of pre-training phases. The pre-training is divided into three stages: full training, selective training, and pruning. In stage one, all model parameters of S-KAN are trained for a specified number of epochs to initially optimize the model parameters and ensure reasonable initial performance. Then, during the weight selective training phase, all parameters in S-KAN except the activation function weights \( w \) are frozen, and only the weights are trained for a specified number of epochs to reorder the importance of the activation functions in the pool. The importance of the activation functions for each node in S-KAN will differ. In the pruning phase, we remove the least significant activation functions from each node to eliminate those that do not contribute significantly to the node's function fitting. The pre-training phase of S-KAN repeats these steps, continuously pruning until only one activation function remains at each node. In this case, Eq\ref{eq:ka} evolves into Eq\ref{eq:kan_model}, and each KAN node will obtain a nonlinear mapping relationship beneficial for reducing the objective function from the selectable activation function pool.
\begin{equation}
f(x_1, x_2, \ldots, x_n) = \sum_{i=1}^{2n+1} \phi_i^{(K_p)} \left( \sum_{j=1}^n \phi_{ij}^{(K_p)} (x_j) \right)
\label{eq:kan_model}
\end{equation}

Due to the extensive use of activation functions in the pre-training phase, S-KAN involves significant computational time and complexity, which limits its applications in some deep networks like CNN. Therefore, we typically use one percent of the overall data for pre-training to improve the efficiency of S-KAN. After pre-training, the entire dataset is used to train the S-KAN with the selected activation function combinations. For multi-class classification problems, stratified sampling is used for data selection, whereas for regression problems, interval sampling is preferred. When fitting binary functions using the $[$$2$ $\rightarrow$ $5$ $\rightarrow$ $1$$]$ framework proposed by the Kolmogorov-Arnold theorem, or fitting slightly more variables such as ternary or quaternary functions, using the full dataset for pre-training has been validated to achieve more precise results within acceptable time costs.

\subsection{Selectable Convolutional KAN (S-ConvKAN)}
The traditional convolutional neural network (CNN), improved from the multilayer perceptron (MLP), can be viewed as a partially connected neural network with shared parameters. It has proven to be highly effective in the field of computer vision, particularly in image recognition and classification \cite{pang2016text,chen2018recurrent,mills2023aio}. Specifically, CNNs set learnable weight parameters in convolutional kernels. The convolutional kernel slides over the image, using these weight parameters to process local regions of the original image, and then generates the final output through a nonlinear activation function. Thus, the convolutional kernel is responsible for linear weighted processing, while nonlinear activation processing occurs after the convolution operation.

Similarly, the Kolmogorov-Arnold Networks, improved and refined by Liu \cite{liu2024kan}, can be modified into convolutional neural networks using a similar approach. The works of Drokin \cite{drokin2024kolmogorovarnoldconvolutionsdesignprinciples} and Bodner \cite{bodner2024convolutional} have also demonstrated the feasibility of this idea. Specifically, in traditional convolutional neural networks, the convolutional kernel only retains linear weights. In the convolutional kernel of ConvKAN, however, each position retains a nonlinear activation function with learnable parameters. These nonlinear activation functions process local regions of the input image or feature map and then directly add up to obtain the convolution result. Bodner \cite{bodner2024convolutional} used the improved B-Spline function proposed by Liu \cite{liu2024kan} as the nonlinear activation function in each convolutional kernel of the ConvKAN.

Eqs \ref{cnn} and \ref{cnnkan} are examples of the traditional CNN kernel and the ConvKAN kernel, both with a size of $2\times2$. In the kernel, \( w \) represents the learnable linear weights, while \( \phi(x) \) represents the learnable B-spline fitting functions in the ConvKAN kernel.

\begin{equation} 
\label{cnn}
\textit{CNN kernel=}
\begin{bmatrix}
w_{1,1} & w_{1,2} \\
w_{2,1} & w_{2,2} \\
\end{bmatrix}
\end{equation} 

\begin{equation} 
\label{cnnkan}
\textit{ConvKAN kernel=}
\begin{bmatrix}
\phi_{1,1}^{B-spline} & \phi_{1,2}^{B-spline} \\
\phi_{2,1}^{B-spline} & \phi_{2,2}^{B-spline}\\
\end{bmatrix}
\end{equation} 

For a local region of the image \( x \) with a size of \( m \times n \):

\[
\begin{bmatrix}
x_{\alpha, \beta} & x_{\alpha, \beta+1} & \cdots & x_{\alpha, \beta+n-1} \\
x_{\alpha+1, \beta} & x_{\alpha+1, \beta+1} & \cdots & x_{\alpha+1, \beta+n-1} \\
\vdots & \vdots & \ddots & \vdots \\
x_{\alpha+m-1, \beta} & x_{\alpha+m-1, \beta+1} & \cdots & x_{\alpha+m-1, \beta+n-1} \\
\end{bmatrix}
\]

The convolution operations using the two types of \( m \times n \) convolution kernels, traditional and ConvKAN, are respectively given by Eqs \ref{cnnout} and \ref{cnnkanout}, where \(\psi\) represents the activation function applied to the convolution result, such as a ReLU or sigmoid function.

\begin{equation} 
\label{cnnout}
\textit{Conv output} = \psi\left(\sum_{i=0}^{m-1}\sum_{j=0}^{n-1} w_{i,j} \cdot x_{\alpha+i,\beta+j} + b\right) 
\end{equation}

\begin{equation} 
\label{cnnkanout}
\textit{ConvKAN output} = \sum_{i=0}^{m-1}\sum_{j=0}^{n-1} \phi_{i,j}\left( x_{\alpha+i,\beta+j}\right)
\end{equation}

It is worth noting that theoretically, any learnable univariate function fitting method can be used as an activation function in ConvKAN. This includes, but is not limited to, Chebyshev polynomials, Jacobi polynomials, radial basis functions (RBF), and wavelet transforms mentioned earlier. Therefore, consistent with the activation function selection strategy applied in the fully connected KAN, these selectable activation function strategies are also applicable to ConvKAN. The figure below shows an example of a selectable KAN convolutional kernel with a size of $2\times2$.

\begin{equation}
\textit{S-ConvKAN kernel=}
\begin{bmatrix}
\phi_{1,1}^{B-spline} & \phi_{1,2}^{Wavelet} \\
\phi_{2,1}^{Jacobi} & \phi_{2,2}^{RBF}\\
\end{bmatrix}
\end{equation}

This diverse activation function selection provides Convolutional KAN with great flexibility and adaptability, allowing it to choose the most suitable combination of nonlinear activation functions based on the specific task requirements, thereby optimizing model performance.

\begin{table*}[h]
\centering
\small
\label{table:comparison}
\fontsize{9pt}{9pt}\selectfont
\setlength{\tabcolsep}{1mm}
\begin{tabular}{@{}lccccccc@{}}
\toprule
Method & $e^{\sin(x_1^2 + x_2^2)}$ & $e^{\sin(\pi x_1) + x_2^2}$ & $e^{J_0(20)x_1 + x_2^2}$ & $\frac{x_1}{x_2}$ & $x_1 \cdot x_2$ & $(x_1 + x_2) + x_1 \cdot x_2$ & $\tanh(5(x_1^4 + x_2^4 + x_3^4 - 1))$ \\ \midrule
KAN & 0.6618 & 403.9234 & 194.7122 & 83.8479 & 1.8517 & 4.5709 & 0.3238 \\ 
FastKAN & 0.3343 & 218.3666 & 69.9517 & 85.7614 & 0.2997 & 0.3435 & 0.1947 \\ 
ChebyKAN & 0.0034 & 19.3789 & 20.0403 & 90.5357 & 0.0016 & 0.0221 & 0.0066 \\ 
Gram & 0.6617 & 403.8912 & 194.6967 & 81.9827 & 1.8518 & 4.5709 & 0.3239 \\ 
WavKAN(Mexican hat) & 0.0349 & 30.6899 & 7.4835 & \underline{49.1050} & 0.0084 & 0.0139 & 0.0369 \\ 
WavKAN(Dog) & 0.0212 & 50.9377 & 9.2506 & 48.1541 & 0.0054 & 0.0121 & 0.0691 \\ 
WavKAN(Shannon) & 0.0317 & 19.8715 & \underline{3.5147} & 71.0755 & 0.0129 & 0.0164 & 0.0332 \\ 
JacobiKAN & 0.6618 & 403.8831 & 194.6982 & 81.9827 & 1.8519 & 4.5708 & 0.3238 \\ 
BernsteinKAN & 0.6618 & 403.8872 & 194.7016 & 81.9827 & 1.8518 & 4.5708 & 0.3236 \\ 
ReLUKAN & 0.0023 & 16.9445 & 28.5015 & 230.3855 & 0.0141 & 0.0296 & \underline{0.0220} \\ 
BottleNeckGram & 0.6618 & 403.8841 & 194.6989 & 81.9827 & 1.8519 & 4.5706 & 0.3318 \\ 
FasterKAN & 0.6632 & 403.1867 & 194.6364 & 84.2065 & 1.8319 & 4.6113 & 0.3318 \\ 
RBFKAN & \underline{0.0025} & \underline{2.6315} & 19.8748 & 97.1193 & \underline{0.0015} & \textbf{0.0116} & 0.0225\\ 
MLP & 0.4307 & 180.3786 & 43.4192 & 80.8309 & 0.0766 & 0.0503 & 0.1855 \\
\textbf{S-KAN(Ours)} & \textbf{0.0023} & \textbf{2.2290} & \textbf{2.3537} & \textbf{39.3020} & \textbf{0.0004} & \underline{0.0119} & \textbf{0.0003} \\ 
 \bottomrule
\end{tabular}

\caption{Performance of KAN and MLP with identical structures in various function fitting tasks. Mean squared error of fitting is used as the evaluation criterion. The best performance is indicated in \textbf{bold}, and the second best is marked with an \underline{underline}.}
\end{table*}

\section{Experiments}

\begin{table*}[h]
\centering
\small
\label{table:performance_comparison}
\fontsize{9pt}{9pt}\selectfont
\begin{tabular}{@{}lccccccc@{}}
\toprule
Method & MNIST(KAN Only) &  MNIST & Fashion MNIST  & CIFAR-10 &CFAR-100  \\ \midrule
KAN &  97.21  &  98.60 / \underline{99.25}$^{\star}$ & 89.77 / 91.16$^{\star}$  & 65.95 / 64.04$^{\star}$ & 41.80 / 43.62$^{\star}$   \\ 
FastKAN & 97.30  &   98.37 / 99.01$^{\star}$ & 88.89 / 90.67$^{\star}$   & 65.10 / 63.01$^{\star}$ & 35.66 / 25.89$^{\star}$ \\ 
ChebyKAN &  96.52  &  \underline{98.87} / - & 90.22 / -  & - / - & - / -  \\ 
Gram &   97.02  &  98.43 / 99.17$^{\star}$ & 90.97 / \underline{91.50}$^{\star}$   & 69.03 / \underline{70.95}$^{\star}$ & 44.39 / 45.49$^{\star}$ \\ 
WavKAN(Mexican hat) & 96.95   &  98.67 / 99.08$^{\star}$ & 90.36 / 90.72$^{\star}$   & 68.29 / 66.32$^{\star}$ & 43.61 / 44.78$^{\star}$ \\ 
WavKAN(Dog) &  96.90  &  98.62 / 99.01$^{\star}$ & 90.24 / 90.53$^{\star}$   & 69.02 / 67.51$^{\star}$ & \underline{46.67} / \underline{47.75}$^{\star}$ \\ 
WavKAN(Shannon) &  96.73  & 98.61 / 99.19$^{\star}$  & 90.13 / 90.44$^{\star}$   & 70.01 / 69.00$^{\star}$ &  45.86 / 46.64$^{\star}$ \\ 
JacobiKAN & 97.13   &  98.83 / 99.21$^{\star}$ & 90.13 / 90.39$^{\star}$   & 70.35 / 68.24$^{\star}$ & 41.92 / 44.20$^{\star}$  \\ 
BernsteinKAN &  97.03  &  97.69 / 98.49$^{\star}$ & 89.64 / 90.45$^{\star}$   & 62.99 / 60.19$^{\star}$ & - / - \\ 
ReLUKAN & 93.19  &  97.84 / 98.15$^{\star}$ & 89.04 / 87.27$^{\star}$   & 58.41 / 52.59$^{\star}$ & 31.87 / 20.31$^{\star}$ \\ 
BottleNeckGram &  \textbf{97.62}  & 98.30 / 99.22$^{\star}$  & \underline{91.20} / 91.15$^{\star}$   & \underline{71.73} / 63.54$^{\star}$ & 46.25 / 47.46$^{\star}$ \\ 
FasterKAN &  96.57   &  98.05 / 98.98$^{\star}$ & 90.22 / 89.59$^{\star}$  & 66.00 / 65.20$^{\star}$ & NA / 36.67$^{\star}$ \\ 
RBFKAN &   97.33  &  98.33 / 98.43$^{\star}$ & 89.53 / 88.71$^{\star}$   & 64.08 / 63.25$^{\star}$ &  35.07 / 22.83$^{\star}$ \\ 
CNN & - & 98.81 &89.98 &69.86 &41.35 \\
\textbf{S-ConvKAN(Ours)} &  \underline{97.41}  &  \textbf{98.94} / \textbf{99.33}$^{\star}$ & \textbf{91.36} / \textbf{92.19}$^{\star}$   & \textbf{72.25} / \textbf{71.50}$^{\star}$  & \textbf{47.52} / \textbf{48.24}$^{\star}$ \\ \bottomrule
\end{tabular}

\caption{Comparison of image classification tasks based on ConvKAN and CNN. "KAN Only" refers to directly flattening the image and using pixels as nodes. Results are presented as "FC / KAN$^{\star}$", where $^{\star}$ indicates the use of KAN as the classifier. Results that were not applicable are marked with "-". S-ConvKAN achieves leading performance in multiple image classification tasks. The best performance is indicated in \textbf{bold}, and the second best is marked with an \underline{underline}. }
\end{table*}

\subsection{Implementation Details}
\subsubsection{Activation Function Pool}
This paper selected a total of 18 activation functions to construct the activation function pool used by S-KAN. These include commonly used activation functions, fitting functions, and polynomials such as B-Spline, Radial Basis Function (RBF), Chebyshev polynomials, GRAM polynomials, Jacobi polynomials, and Bernstein polynomials. The pool also includes various wavelet transforms with different wavelet functions, as well as specially designed activation functions used in FastKAN \cite{li2024kolmogorovarnold}, FasterKAN \cite{fasterkan}, ReLUKAN \cite{qiu2024relu}, and BottleNeckGramKAN \cite{drokin2024kolmogorovarnoldconvolutionsdesignprinciples}.

\subsubsection{Fitting}
We evaluated the performance of S-KAN on binary and ternary function fitting tasks. The functions were selected based on the examples used by Liu \cite{liu2024kan} with some modifications. For binary functions, we adopted the $[2$ $\rightarrow$ $5$ $\rightarrow$ $1]$ network architecture proposed by the Kolmogorov-Arnold theorem for fitting training. For ternary functions, a variant architecture of $[3$ $\rightarrow$ $5$ $\rightarrow$ $1]$ was used for training. In the function fitting tasks, we used the complete dataset for pre-training, with 20 epochs for both full training and weight training. Pre-training stops when S-KAN completes the activation function selection for each node. Subsequently, the model continues training on the full dataset, with a total of 1000 epochs, including the pre-training phase. For comparison, other KANs with fixed activation functions and MLPs of the same specifications were trained on the full dataset for 1000 epochs. All networks were trained using the Adam optimizer with a learning rate of 0.001.

\subsubsection{Classifaction}
This study used four representative image classification datasets: MNIST \cite{lecun1998gradient}, Fashion MNIST \cite{xiao2017fashion}, CIFAR-10 \cite{krizhevsky2009learning}, and CIFAR-100 \cite{krizhevsky2009learning} to train and test S-KAN and S-ConvKAN. For the MNIST, Fashion MNIST, and CIFAR-10 datasets, we used neural networks constructed with two convolutional layers (kernel size of 5 or 3) followed by max-pooling layers of similar specifications. For CIFAR-100, we used three convolutional layers followed by max pooling. After the convolutional and pooling layers, we used traditional fully connected layers and KAN layers as classification heads and compared their performance differences. Additionally, we tested the performance of fully connected KAN on the MNIST.

Due to the large size of the image classification datasets and the greater complexity of classification networks compared to function fitting tasks, we used only 1\% of the full dataset for pre-training in the S-KAN pre-training phase. Each full training and weight training epoch in the pre-training phase was set to 3. To ensure a fair comparison of models, after the S-KAN pre-training phase, where each node's activation function is selected, we re-initialized the network parameters and trained from scratch. We trained the networks for 30 epochs on the MNIST and Fashion MNIST datasets and 15 epochs on the CIFAR-10 and CIFAR-100 datasets. Similar to the function fitting tasks, we trained convolutional KANs with fixed activation functions and CNNs of the same specifications. When using KAN as the classification head, the activation functions used in each KAN classification head matched the types used in the convolutional KAN. For each network, the Adam optimizer and a learning rate of 0.001 were used for training. All experiments were conducted using NVIDIA Tesla V100 (32G) GPUs for training. The implementation of S-KAN and S-ConvKAN was completed using Torch 2.1.0. All the specific structures are provided in the appendix.
\begin{figure*}[htbp]
\centering
\includegraphics[width=0.9\textwidth]{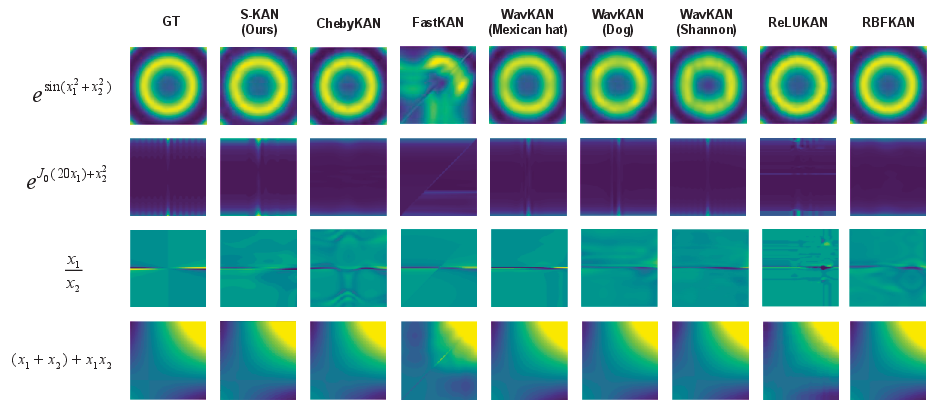}
\caption{Fitting of four binary functions. Visualization shows that the S-KAN can achieve high-performance fitting across different binary functions (exponential sum of squares, product, division), indicating its universality.}
\label{fitting}
\end{figure*}

\subsection{Comparations}
In the fitting tasks, the proposed method was compared with fully connected MLPs with the same number of nodes and KAN structures, as shown in Table 1. For seven typical fitting tasks, S-KAN performed best in four tasks and demonstrated strong general performance. In Figure \ref{fitting}, we visualized the comparison between the fittings of four binary functions and the ground truth. It is evident that S-KAN exhibits high robustness and general fitting capability both quantitatively and qualitatively, far surpassing MLP methods with the same number of nodes and depth.

\begin{figure}[h]
\centering
\includegraphics[width=0.48\textwidth]{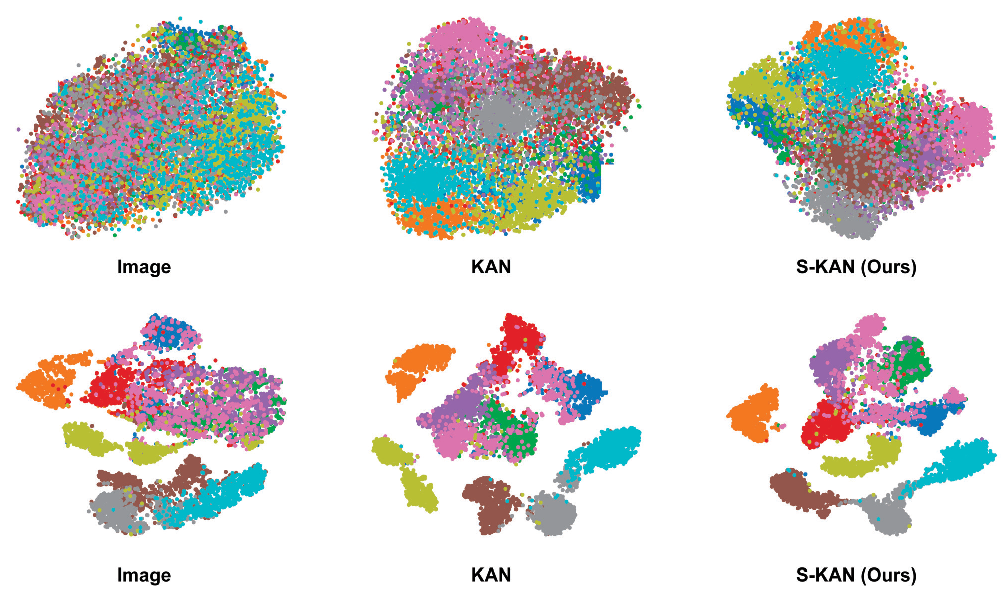}
\caption{Based on cross entropy, t-SNE visualization results for two image datasets with 10 classes each (CIFAR-10 on top and Fashion MNIST on bottom). "Image" represents the embedding of the original image. "KAN" and "S-KAN" represent the visualization results of the penultimate layer's embeddings of the models. The embedding results based on S-KAN show significantly better feature clustering capability compared to the original KAN.}
\label{tsne}
\end{figure}

In image classification, we fairly compared traditional CNNs and known KAN-based Conv-KANs using the same feature extraction and classification network, as shown in Table 2. We compared the use of KAN in both feature extraction and classification heads. S-KAN performed best in image classification using KAN as a classifier, achieving leading performance in four image datasets. In Figure \ref{tsne}, we visualized the feature visualization results of the original KAN without activation pool selection and the proposed S-KAN with activation space selection in Fashion MNIST and CIFAR-10. The adaptive activation function selection and pruning in S-KAN enable better fitting of the original data distribution and better generalization of different class biases.

\section{Conclusion \& Discussion}
By examining the recently proposed KAN, it is feasible to achieve efficient function fitting by relying on nonlinear activation. Building on this insight, we propose an S-KAN with activation space selection capability. It integrates the K-A theorem and uses weighted activation functions for each data node, implementing pruning to select the nonlinear mappings most beneficial for reducing the objective function. Based on this approach, we also improved CNN and introduced S-ConvKAN as a feature extractor for image classification. We conducted extensive tests and comparisons on the basic tasks of the two machine learning paradigms, to illustrate the effectiveness of our proposed methods. In the fitting tasks, S-KAN exhibited outstanding general performance in seven typical function fitting scenarios. For general computer vision image classification tasks, S-ConvKAN also demonstrated superior performance, surpassing CNN methods with the same structure and KAN-based visual strategies. In summary, S-KAN changes the traditional approach of designing KANs with a single activation function. By constructing an activation space and selecting activation functions that are beneficial for establishing reliable data relationships, S-KAN significantly enhances the performance of networks, by using learnable nonlinear activation for function fitting and feature extraction. It is suitable for broader applications in areas such as large language models, deep vision models, and multimodal models.

\subsubsection{Limiatation}
Replacing linear computations with nonlinear ones has led to increased training time, representing a limitation of KAN-based designs. Although this study attempts to expedite the selection of the optimal activation function by pre-training with just 1\% of the samples, most existing accelerated computations are designed for linear methods, leaving substantial room for improvement in the efficiency of computations for S-KAN.

\subsubsection{Future Work}
This paper introduces S-KAN from the perspective of selectivity in the activation space, but restricts the selective operations to data nodes rather than hidden layer nodes. A KAN that is selective at the hidden layer, could be a potentially effective paradigm, but at a cost of computational complexity. Developing parallel and efficient KAN methods, along with dedicated hardware acceleration techniques, is expected to relieve this problem in the future.

\bibliography{aaai2025}
\clearpage
\section*{Appendix}

\subsection{Selection of Activation Functions in S-KAN for Function Fitting Examples}

\begin{figure*}[h]
    \centering
    \includegraphics[scale=0.8]{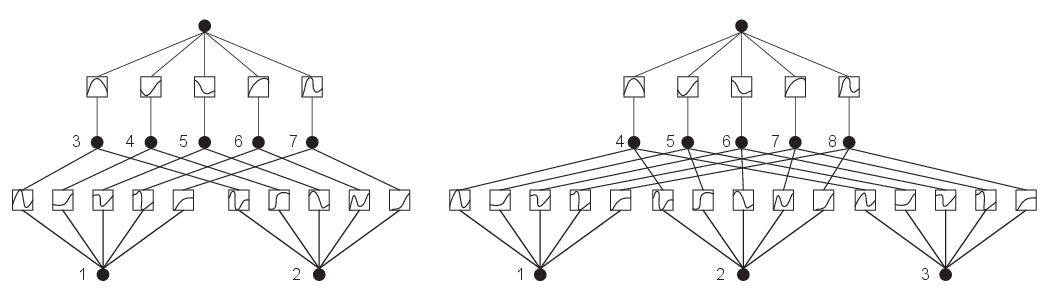}
    \caption{Node positions in the network structures used for S-KAN binary (left) and ternary (right) regression tasks.}
    \label{fig}
\end{figure*}

In the experiments, we used seven example functions to test the selectable KAN in regression tasks. For the binary functions, we used a [2→5→1] structure for fitting, while for the ternary function examples, we used a [3→5→1] structure, as illustrated in Figure \ref{fig}.

As shown in Table \ref{ac_sele}, for each example function, the selectable KAN effectively chose the appropriate activation functions, successfully completing various fitting tasks.

\begin{table*}[h]
\centering
\fontsize{9pt}{9pt}\selectfont
\setlength{\tabcolsep}{1mm}
\begin{tabular}{cccccccc}
\toprule
\textbf{Node} & \textbf{$e^{\sin(x_1^2 + x_2^2)}$} & \textbf{$e^{\sin(\pi x_1) + x_2^2}$} & \textbf{$e^{J_0(20)x_1 + x_2^2}$} & \textbf{$\frac{x_1}{x_2}$} & \textbf{$x_1 \cdot x_2$} & \textbf{$(x_1 + x_2) + x_1 \cdot x_2$} & \textbf{$\tanh(5(x_1^4 + x_2^4 + x_3^4 - 1))$} \\ 
\midrule
1 & RBF & RBF & ReLU & FastKAN & Chebyshev & RBF & RBF \\ 
2 & Chebyshev & Chebyshev & Chebyshev & Wavelet & RBF & Chebyshev & Chebyshev \\ 
3 & RBF & RBF & Chebyshev & ReLU & FastKAN & Wavelet & Chebyshev \\ 
4 & Chebyshev & Chebyshev & Wavelet & RBF & RBF & Chebyshev & RBF \\ 
5 & Chebyshev & RBF & Wavelet & Wavelet & Chebyshev & FastKAN & RBF \\ 
6 & RBF & RBF & Wavelet & Wavelet & RBF & RBF & RBF \\ 
7 & Chebyshev & RBF & Wavelet & Wavelet & Chebyshev & Chebyshev & Chebyshev \\ 
8 &  &  &  &  &  &  & Chebyshev \\ 
\bottomrule
\end{tabular}

\caption{Selected activation functions for each node in S-KAN across different example regression tasks.}
\label{ac_sele}
\end{table*}

\subsection{Network Configuration of ConvKAN}

To ensure fairness in the experiments, we standardized the network depth and configuration across each image classification dataset. Here, \textit{ConvKAN} refers to the convolutional form of KAN. \textit{ConvKAN(a→b, m$\times$m)} denotes a convolutional KAN with an input channel of ‘$a$’, an output channel of ‘$b$’, and an $m\times m$ convolutional kernel. \textit{FC / KAN (n)} represents a fully connected (\textit{nn.Linear}) layer or a regular KAN layer with an output dimension of ‘$n$’. In the experiments, we compare S-ConvKAN with ConvKAN using different activation functions and a conventional CNN of similar structure. The specific network configurations are detailed in the table below.

\begin{table*}[h]
\centering
\fontsize{9pt}{9pt}\selectfont
\setlength{\tabcolsep}{1mm}
\begin{tabular}{ccccc}
\toprule
\textbf{Layer} & \textbf{MNIST} & \textbf{Fashion MNIST} & \textbf{CIFAR-10} & \textbf{CIFAR-100} \\ 
\midrule
1 & ConvKAN(1→10, 5$\times$5) & ConvKAN(1→32, 3$\times$3) & ConvKAN(3→32, 3$\times$3) & ConvKAN(3→32, 3$\times$3) \\ 
2 & maxpool(2$\times$2) & maxpool(2$\times$2) & maxpool(2$\times$2) & maxpool(2$\times$2) \\ 
3 & ConvKAN(10→20, 5$\times$5) & ConvKAN(32→64, 3$\times$3) & ConvKAN(32→16, 3$\times$3) & ConvKAN(32→64, 3$\times$3) \\ 
4 & maxpool(2$\times$2) & maxpool(2$\times$2) & maxpool(2$\times$2) & maxpool(2$\times$2) \\ 
5 & flatten & flatten & flatten & ConvKAN(64→128, 3$\times$3) \\ 
6 & FC / KAN(50) & FC / KAN(50) & FC / KAN(50) & maxpool(2$\times$2) \\ 
7 & FC / KAN(10) & FC / KAN(10) & FC / KAN(10) & flatten \\ 
8 &  &  &  & FC / KAN(256) \\ 
9 &  &  &  & FC / KAN(100) \\ 
\bottomrule
\end{tabular}

\caption{Network configurations of ConvKAN for different image classification datasets.}
\label{network_setting}
\end{table*}

\subsection{Parameter Comparison}

We conducted a parameter comparison of the networks used in the experiments. The parameter count for S-KAN represents the final model’s parameter count, i.e., the parameter count after the activation function selection. As shown in Table \ref{param}, in each cell, the number on the left indicates the parameter count when using nn.Linear as the classifier, while the number on the right indicates the parameter count when using KAN as the classifier. For the binary function regression tasks, the activation functions selected by S-KAN vary for each task, leading to different parameter counts; therefore, these are not specified.

\begin{table*}[h]
\centering
\fontsize{9pt}{9pt}\selectfont
\begin{tabular}{cccccc}
\toprule
& \textbf{Binary / Ternary Regression}  & \textbf{MNIST} & \textbf{Fashion MNIST} & \textbf{CIFAR-10} & \textbf{CIFAR-100} \\ 
\midrule
KAN & 149  / 194 & 64K / 195K$^{\star}$ & 65K / 228K$^{\star}$ & 73K / 308K$^{\star}$ & 1.4M / 5.8M$^{\star}$ \\ 
FastKAN & 155 / 202 & 64K / 197K$^{\star}$ & 65K / 230K$^{\star}$ & 74K / 309K$^{\star}$ & 1.4M / 5.8M$^{\star}$ \\ 
ChebyKAN & 75 / 125 & 42K / 108K$^{\star}$ & 45K / 127K$^{\star}$ & 54K / 171K$^{\star}$ & 1M / 3.2M$^{\star}$ \\ 
Gram & 95 / 120 & 42K / 108K$^{\star}$ & 45K / 127K$^{\star}$ & 54K / 171K$^{\star}$ & 1M / 3.2M$^{\star}$ \\ 
WavKAN & 72 / 92 & 37K / 87K$^{\star}$ & 40K / 102K$^{\star}$ & 49K / 137K$^{\star}$ & 0.9M / 2.6M$^{\star}$ \\ 
JacobiKAN & 102 / 132 & 48K / 130K$^{\star}$ & 50K / 152K$^{\star}$ & 59K / 205K$^{\star}$ & 1.1M / 3.9M$^{\star}$ \\ 
BernsteinKAN & 102 / 132 & 48K / 130K$^{\star}$ & 50K / 152K$^{\star}$ & 59K / 205K$^{\star}$ & 1.1M / 3.9M$^{\star}$ \\ 
ReLUKAN & 238 / 294 & 63K / 184K$^{\star}$ & 64K / 228K$^{\star}$ & 73K / 288K$^{\star}$ & 1.3M / 5.2M$^{\star}$ \\ 
BottleNeckGram & 84 / 120 & 35K / 57K$^{\star}$ & 42K / 76K$^{\star}$ & 51K / 118K$^{\star}$ & 0.7M / 1.6M$^{\star}$ \\ 
FasterKAN & 134 / 176 & 59K / 175K$^{\star}$ & 60K / 204K$^{\star}$ & 69K / 275K$^{\star}$ & 1.3M / 5.1M$^{\star}$ \\ 
RBFKAN & 141 / 186 & 64K / 196K$^{\star}$ & 64K / 215K$^{\star}$ & 73K / 308K$^{\star}$ & 1.4M / 5.8M$^{\star}$ \\ 
S-KAN & - / 148 & 46K / 96K$^{\star}$ & 47K / 122K$^{\star}$ & 56K / 177K$^{\star}$ & 1M / 2.6M$^{\star}$ \\ 
MLP / CNN & 21 / 26 & 22K & 26K & 34K & 0.6M \\ 
\bottomrule
\end{tabular}

\caption{Parameter comparison across different networks and tasks. Results are presented as "FC / KAN$^{\star}$", where $^{\star}$ indicates the use of KAN as the classifier.}
\label{param}
\end{table*}

\subsection{Supplementary Experiments on S-KAN vs. MLP and CNN}

As shown in Table \ref{param}, the parameter count of S-KAN is significantly higher than that of MLP / CNN with the same structure. To study the performance of S-KAN compared to MLP / CNN with similar parameter counts, we widened the MLP / CNN networks used in the previous experiments while maintaining the existing depth. For example, in the binary regression tasks, we used a [2→30→1] structure for function regression; in the image classification tasks, we increased the output channels of the existing CNN to match the parameter count of S-KAN. As shown in the Table \ref{param1}, “MLP / CNN complex” refers to these traditional neural networks with increased complexity.

We tested these networks on the existing tasks, and the results are shown in Table \ref{ac_sele} and Table \ref{extra2}. It can be seen that, except for the extremely simple binary regression task, such as $f(x_1, x_2) = x_1\times x_2$, the overall performance of MLP / CNN with similar parameter counts still falls short of S-KAN.

\subsection{Pseudo Code of S-KAN}
\begin{algorithm}
\caption{SelectableKANLayer: Training and Pruning}
\begin{algorithmic}[1]
\small
\State \textbf{Class} SelectableKANLayer:
\State \quad \textbf{def} \_\_init\_\_(self):
\State \quad \quad Initialize base setup and KAN layer factory
\State \quad \quad Create layer structure for each input dimension
\State \quad \quad Initialize attention weights 
\State \quad \textbf{def} forward(self, x):
\State \quad \quad \textbf{for} each input feature \textbf{in} x:
\State \quad \quad \quad Compute outputs for all KAN layers
\State \quad \quad \quad Apply attention weights for weighted sum
\State \quad \quad Combine outputs for all input features to compute final output
\State \quad \textbf{def} prune(self):
\State \quad \quad \textbf{for} each input feature:
\State \quad \quad \quad If more than one layer exists, remove the least important layer
\State \quad \quad \quad Update attention weights
\State \quad \quad \quad If only one layer exists, maintain current weights
\State 
\State \textbf{Training Strategy:}
\State \quad \textbf{Step 1: Full Training}
\State \quad \quad Use all candidate nonlinear mapping methods
\State \quad \quad Initialize and update weights to minimize the cost function
\State \quad \textbf{Step 2: Selective Training}
\State \quad \quad Train only selected weights to optimize the cost function
\State \quad \textbf{Step 3: Pruning}
\State \quad \quad Prune unimportant activation functions
\State \quad \quad Reallocate remaining weights to ensure their sum is 1
\end{algorithmic}
\end{algorithm}

\begin{table*}[h]
\centering
\fontsize{9pt}{9pt}\selectfont
\begin{tabular}{cccccc}
\toprule
 & \textbf{Binary / Ternary Regression} & \textbf{MNIST} & \textbf{Fashion MNIST} & \textbf{CIFAR-10} & \textbf{CIFAR-100} \\ 
\midrule
S-KAN & - / 148 & 46K & 47K & 56K & 1M \\ 
MLP / CNN & 21 / 26 & 22K & 26K & 34K & 0.6M \\ 
MLP / CNN complex & 121 / 151 & 53K & 54K & 60K & 1M \\ 
\bottomrule
\end{tabular}

\caption{Parameter comparison of S-KAN, MLP / CNN, and MLP / CNN with increased complexity across different tasks.}
\label{param1}
\end{table*}

\begin{table*}[h]
\centering
\fontsize{9pt}{9pt}\selectfont
\setlength{\tabcolsep}{0.5mm}
\begin{tabular}{cccccccc}
\toprule
 & \textbf{$e^{\sin(x_1^2 + x_2^2)}$} & \textbf{$e^{\sin(\pi x_1) + x_2^2}$} & \textbf{$e^{J_0(20)x_1 + x_2^2}$} & \textbf{$\frac{x_1}{x_2}$} & \textbf{$x_1 \cdot x_2$} & \textbf{$(x_1 + x_2) + x_1 \cdot x_2$} & \textbf{$\tanh(5(x_1^4 + x_2^4 + x_3^4 - 1))$} \\ 
\midrule
S-KAN & 0.0023 & 2.2290 & 2.3537 & 39.3020 & 0.0004 & 0.0119 & 0.0003 \\ 
MLP & 0.4307 & 180.3786 & 43.4192 & 80.8309 & 0.0766 & 0.0503 & 0.1855 \\ 
MLP complex & 0.1169 & 70.5626 & 15.7897 & 60.6777 & 0.0003 & 0.0080 & 0.0764 \\ 
\bottomrule
\end{tabular}

\caption{Selected activation functions for each node in S-KAN across different example regression tasks.}
\label{ac_sele}
\end{table*}

\begin{table*}[!h]
\centering
\fontsize{9pt}{9pt}\selectfont

\begin{tabular}{cccccccc}
\toprule
 & \textbf{MNIST} & \textbf{Fashion MNIST} & \textbf{CIFAR-10} & \textbf{CIFAR-100} \\ 
\midrule
S-ConvKAN & 98.94 & 91.36 & 72.25 & 47.52 \\ 
CNN & 98.81 & 89.98 & 69.86 & 41.35 \\ 
CNN complex & 98.90 & 90.24 & 69.88 & 43.16 \\ 
\bottomrule
\end{tabular}
\caption{Accuracy of S-ConvKAN and CNN on different image classification tasks.}
\label{extra2}
\end{table*}

\end{document}